\pdfoutput=1

\documentclass[11pt]{article}

\usepackage{emnlp2021}
\usepackage{times}
\usepackage{latexsym}
\usepackage[T1]{fontenc}
\usepackage[utf8]{inputenc}
\usepackage{microtype}

\usepackage{graphicx}
\usepackage{amsmath}
\usepackage{amssymb}
\usepackage{amsthm}
\usepackage{numprint}
\usepackage{centernot}
\usepackage{booktabs}
\usepackage{multirow}
\usepackage{enumitem}
\usepackage{bbm}
\usepackage{todonotes}
\usepackage{pgf}
\usepackage{array}
\newcolumntype{H}{>{\setbox0=\hbox\bgroup}c<{\egroup}@{}}

\DeclareUnicodeCharacter{2212}{-}

\usepackage{subfiles}

\newcommand{\ra}[1]{\renewcommand{\arraystretch}{#1}}

\newcommand{\lexprobing}{Overlap}
\newcommand{\subprobing}{Subsequence}
\newcommand{\subprobingcond}{Sub}
\newcommand{\negprobing}{NegWords}

\newcommand{\etoe}{\textsubscript{e2e}}

\newcommand{\Lon}[0]{L_{\mathrm{online}}}
\newcommand{\Lunif}[0]{L_{\mathrm{unif}}}

\title{Debiasing Methods in Natural Language Understanding \\ Make Bias More Accessible}

\author{
    ~~Michael Mendelson
    \hspace{10em}
    Yonatan Belinkov\thanks{~~Supported by the Viterbi Fellowship in the Center for Computer Engineering at the Technion.} 
    \\
    \texttt{michael.me@cs.technion.ac.il}
    \hspace{2em} 
    \texttt{belinkov@technion.ac.il}
    \vspace{0.3em} \\
    The Henry and Marilyn Taub Faculty of Computer Science \\
    Technion -- Israel Institute of Technology
 }

\date{}

\begin{document}
\maketitle
\begin{abstract}
Model robustness to bias is often determined by the generalization on carefully designed out-of-distribution datasets. Recent debiasing methods in natural language understanding (NLU) improve performance on such datasets by pressuring models into making unbiased predictions. An underlying assumption behind such methods is that this also leads to the discovery of more robust features in the model’s inner representations. We propose a general probing-based framework that allows for post-hoc interpretation of biases in language models, and use an information-theoretic approach to measure the extractability of certain biases from the model's representations. We experiment with several NLU datasets and known biases, and show that, counter-intuitively, the more a language model is pushed towards a debiased regime, the more bias is actually encoded in its inner representations.\footnote{Our code and data are available at: \url{https://github.com/technion-cs-nlp/bias-probing}.}

\end{abstract}

\section{Introduction}
\label{sec:intro}
\begin{figure}
    \centering
    \includegraphics[width=\linewidth]{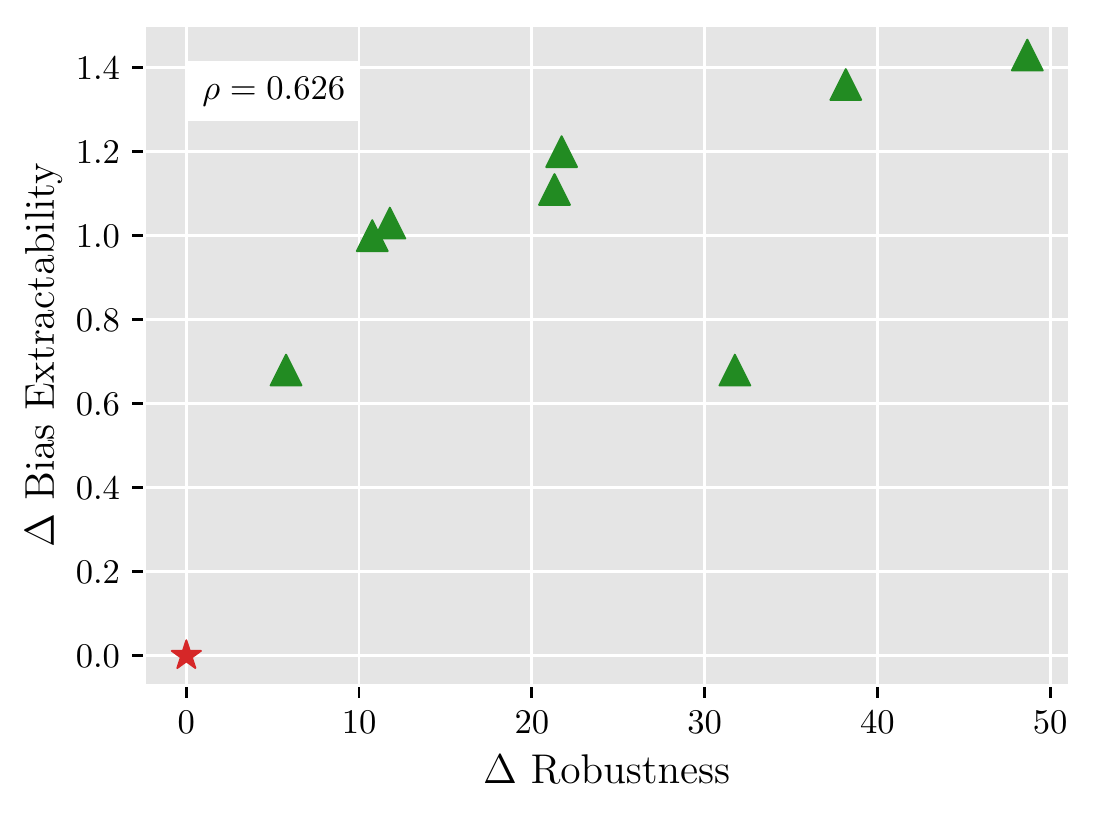}
    \caption{Amount of subsequence bias extracted from different language models vs.\  the robustness of models to the bias. Robustness is measured as improvement of the model on out-of-distribution examples, while extractability is measured as the improvement of the probe's ability to extract the bias from a debiased model, compared to the baseline.}
    \label{fig:mnli_lex_class_scatter}
\end{figure}
State of the art neural language models such as BERT \citep{devlin2018bert} usually work by pre-training an encoder to learn universal word representations, and then fine-tuning it on some classification or regression task.  
From a robustness point of view, such pretrain-and-fine-tune pipelines are known to be prone to biases that are present in data \citep{gururangan-etal-2018-annotation,poliak-etal-2018-hypothesis, mccoy2019right, schuster2019debiasing}. Various methods were proposed to mitigate such biases in a form of robust training, where a \textit{bias model} is trained to capture the bias and then used to relax the predictions of a main model, so that it can focus less on biased examples and more on the ``hard'', more challenging examples \citep[][inter alia]{clark2019dont,mahabadi2020endtoend,utama2020debiasing, sanh2020learning}.
Then, the resulting model is evaluated on out-of-distribution (o.o.d) data, in the form of challenge datasets containing ``hard'' examples that were deliberately constructed to be anti-biased. Examples of such datasets include HANS \citep{mccoy2019right} for natural language inference (NLI) and FEVER-Symmetric \citep{schuster2019debiasing} for fact verification. An underlying assumption behind this methodology is that better generalization out of distribution also means that the model learned more robust features. However, while evaluation using challenge datasets only relays information about the generalization of the model through predictions, it does not reveal what actually caused it and how the internal representations were affected.

To assess whether bias has been removed from the internal representations, we design probing tasks targeting several known biases: lexical overlap biases and negative word bias. While probing is usually concerned with simple linguistic properties such as part-of-speech tags \citep{belinkov:2019:tacl},  we instead define probing tasks with the purpose of revealing bias in the representations. An example of such probing task is to predict whether a sentence-pair is lexically overlapping given only access to their joint representation---a classifier which is able to label the pair by this property consequently must use information about the bias which is encoded in the representation. We construct probing datasets for assessing bias in several natural language understanding (NLU) datasets. Lastly, we use information-theoretic probing  \citep{voita2020informationtheoretic} to analyze the extractability of bias from vanilla and debiased models using the probing classifier.

We conduct experiments on two NLI datasets and one fact verification dataset across a variety of debiasing methods and bias types, and test whether the bias removal is as successful as o.o.d evaluation suggests. Surprisingly, we discover that making models robust from the perspective of the down-stream task, causes the inner representations to encode more of the information about the specific bias in question. Figure~\ref{fig:mnli_lex_class_scatter} shows an example of this trend in NLI, where as robustness of the fine-tuned model to biased predictions increases, so does the ability of the probing classifier to extract bias. 

To summarize, we make the following contributions:
\begin{itemize}[itemsep=2pt,topsep=2pt,parsep=2pt]
    \item We present a general probing-based framework to measure extractability of bias from inner model representations.
    \item We use this framework to construct several new probing tasks based on well-studied dataset biases in NLU tasks.
    \item We show that pressuring a model into making unbiased predictions actually makes biased features more extractable from the model representations.
\end{itemize}

\section{Related Work}
\label{sec:related_work}
\subsection{Dataset Biases}

Deep neural models are prone to shortcut learning \citep{geirhos2020shortcut}, by discovering and using idiosyncratic biases, heuristics, and statistical cues in the data. For example, \citet{poliak-etal-2018-hypothesis} showed that the Stanford natural language inference dataset (SNLI; \citealt{snli:emnlp2015}) contains ``give-away'' words, i.e., words $w$ which have a high value of $p \left( l \mid w \right)$ w.r.t a given label $l$. They noticed that 4 out of the 10 words with the highest $p \left( \mathrm{contradiction} \mid w \right)$ are universal negation words,\footnote{nobody, alone, no, empty.} suggesting that negation is strongly correlated with contradiction in the data. These clues appeared in the hypothesis side, making them a kind of hypothesis-only bias, where a classifier receiving as input only the hypothesis is able to correctly predict the label \citep{poliak-etal-2018-hypothesis,gururangan-etal-2018-annotation}. A similar type of bias, known as claim-only bias, is found in the FEVER fact verification dataset \citep{thorne-etal-2018-fever}, and was also associated with a strong correlation of negation words with the labels in the dataset \citep{schuster2019debiasing}.  
Another kind of bias is the association of entailment with cases of lexical overlap between the premise and hypothesis. This bias leads to poor performance of models on the HANS challenge dataset \citep{mccoy2019right}, where all samples contain lexical overlap and non-entailed samples are formed such that the bias does not entail the label. 
This suggests that models rely on features that are cues for lexical overlap bias when predicting the entailment of premise--hypothesis pairs.

\subsection{Bias Mitigation and Robustness}
\label{sec:bias_mitigation}

Recent work on bias mitigation attempts to create more robust models by training a combination model, based on the main model. The main model, parameterized by $\theta_m$, is a non-robust language model. The bias model, parameterized by $\theta_b$, is a \textit{weak model} whose purpose is to model the biases during training, by minimizing a loss $\mathcal{L}_b$. The objective of the combination model is to minimize a combined loss function
$\mathcal{L}_c \left( \theta_m, \theta_b \right)$, such that the main model leverages knowledge about bias in data, obtained using the weak model. This pipeline is general, and it allows models to be trained either end-to-end, or step-by-step by first training the bias model and then using its predictions to robustly train the main model. Recent papers show that such techniques are effective when evaluated on challenge datasets specifically designed to target known biases and hard examples \citep{he2019unlearn, clark2019dont,utama2020debiasing, UtamaDebias2020degrading, sanh2020learning, mahabadi2020endtoend}. However, this approach does not ensure that the model indeed learns more robust features, nor does it shed light on exactly \emph{how} the feature detectors react to this change, and how the bias is represented in the model.

\subsection{Probing}
Probing was somewhat successfully used to analyze sentence embeddings and to show that such models capture surface features such as sentence length, word content, and the order of words \citep{adi2017finegrained}, or various syntactic and semantic features \citep{conneau-etal-2018-cram}; see \citet{belinkov:2019:tacl} for a survey. In contrast, we focus our analysis on \textit{biased} features, and employ advances in probing methodology to analyze two kinds of bias---lexical overlap and negation bias. Designing probes to accurately interpret the desired behavior is not trivial and measuring their accuracy is insufficient, since the probing classifiers are prone to memorization and bias as well \cite{hewitt-liang-2019-designing}, among other shortcomings \cite{belinkov2021probing}. 
Recently, \citet{voita2020informationtheoretic} presented an information-theoretic approach for evaluating probing classifiers, which  accounts for the complexity of the probing classifier by measuring its minimum description length (MDL).  MDL measures how efficiently a model can extract information about the labels from the inputs, and we use it as a measure of extractability of certain biases from model representations.

\section{Methods}
\label{sec:methods}
We lay down a general framework for interpreting bias in inner model representations.
Given a model $f_\theta:X\rightarrow Y$ with learnable parameters $\theta$,
we assume that it can be decoupled into two stages:
\begin{itemize}
    \item A representation layer (or multiple layers) with learnable parameters $\theta_1$, which we denote $\mathcal{R}_{\theta_1} : X \rightarrow Z$, maps samples from the input space to a latent space $Z$, the ``representation''.
    \item A classification layer with learnable parameters $\theta_2$, which we denote $\mathcal{F}_{\theta_2} : Z \rightarrow Y$, maps the latent representations to the final output.
\end{itemize}
We can thus re-define our classifier as
\begin{equation}
     f_{\theta} \left( x \right) \triangleq \mathcal{F}_{\theta_2} \left( \mathcal{R}_{\theta_1} \left( x \right) \right).
\end{equation}

For example, in NLI we assume that data samples are given as sentence pairs $x = \left( p, h \right)$ where $p$ is a premise and $h$ is a hypothesis. $\mathcal{R} \left( p, h \right)$ is the joint representation of the two, and this representation is then used by $\mathcal{F}$ to produce a prediction.

In this work, we compare baseline models fine-tuned on some down-stream task to models debiased during the fine-tuning step. We produce representations from both types of models and measure the extractability of bias using a probing classifier. Our probing tasks are defined in terms of ``bias-revealing'' properties, which are based on a-priori knowledge of the bias in question, and are able to distinguish between biased and unbiased samples from the original dataset. We next describe how to construct such probing tasks and appropriate datasets.

\subsection{Probing Tasks and Datasets}
\label{sec:meth:probing}

We define a probing classifier as a classifier $g_{\Psi} : Z \rightarrow Y_P$ with learnable parameters $\Psi$,  which maps inputs from a latent representation space $Z$ to a probing property space $Y_P$, where $P : X \rightarrow Y_P$ is some real property of the original input, which we call the \textit{probing property}.
Next, we define a \emph{probing dataset} for each probing task:
\begin{equation}
\mathcal{D}_P = \left\{ 
    \left(
        \mathcal{R}_{\theta} \left( x \right), P \left( x \right)
    \right) \mid x \in X
\right\}.
\end{equation}
Lastly, we train the probing classifier on the constructed dataset and evaluate its performance on the probing task. 
We introduce two new probing tasks that target the well researched types of bias present in several datasets: lexical bias and negative word bias. For presentation purposes, consider the NLI task, where data samples are given as sentence pairs $x = \left( p, h \right)$ where $p$ is a premise and \emph{h} is a hypothesis. The extension to fact verification and other pair relationship classification tasks is straightforward.

\paragraph{\negprobing{}}
To analyze negation bias in NLI and fact verification,  
we define a list of negative words $V$\footnote{In our experiments we use $V = \{$no, not, nobody, never, nothing, none, empty, neither, cannot$\} \cup \{$Words that end with \textit{n't}$\} $ for a total of $\lvert V \rvert = 27$ words.} and a sentence pair property
\begin{equation}
    P_{\mathrm{neg}}^{V} \left(p, h \right) = \mathbbm{1} \left[ V \cap h \ne \emptyset \right].
\end{equation}
That is, an example is positive if its hypothesis (in the case of NLI) or claim (in the case of fact verification) contains at least one negative word from the list. This method poses some limitations: For example, we do not consider double negatives in the hypothesis that affect its meaning, or the presence of negation in both premise and hypothesis. However, our construction is consistent with prior findings on negation bias \citep{gururangan-etal-2018-annotation,poliak-etal-2018-hypothesis,schuster2019debiasing}.

\paragraph{\lexprobing{}/\subprobing{}}
Based on the analysis of \citet{mccoy2019right}, we define a class of probing tasks for identifying the different lexical heuristics in NLI. We focus on lexical overlap and subsequences\footnote{We exclude the constituency heuristic since it is not frequent enough in MNLI to construct a probing dataset.} and define two sentence pair properties:
\begin{equation}
    P_\text{lex} (p, h) = \mathbbm{1} \left[ h \subseteq p \right],
\end{equation}
where an example is positive if all the hypothesis words are found in the premise (regardless of word order), 
and
\begin{equation}
P_\text{sub} (p, h) = \mathbbm{1} \left[ h \text{ is a subsequence of } p \right],
\end{equation}
where an example is positive if the hypothesis is a subsequence of the premise. 

\subsection{Data Processing}
\label{sec:data_processing}
To alleviate issues of data balancing, we take the following steps when processing the probing datasets: First, we identify all the biased samples in a given dataset, according to the probing property. Since in all our datasets the positive class (biased samples) is the minority class, we subsample the same amount of samples from the remaining subset (the majority class). We end up with a balanced probing dataset. This ensures that when splitting the data during online code training, and when measuring performance on the entire dataset, the process is unaffected by the bias \textit{evidence}, that is, the amount of bias in the original dataset. 

The probing datasets are constructed from three base NLU datasets: SNLI \citep{snli:emnlp2015}, MultiNLI \citep{williams-etal-2018-broad} and FEVER \citep{thorne-etal-2018-fever}, following the original train/validation/test splits.\footnote{FEVER does not provide a test set, and we therefore report results on the validation set, and do not perform any type of hyperparameter tuning.
} 
Inspired by previous work on biases in NLU datasets (Section~\ref{sec:related_work}), we construct \textbf{\negprobing{}} probing datasets from all three base NLU datasets and \textbf{\lexprobing}/\textbf{\subprobing} probing datasets from SNLI and MNLI.
The dataset statistics are presented in Table \ref{tab:stats_probing_tasks}.

\begin{table}[ht]
    \centering
    \setlength{\tabcolsep}{4pt}
    \begin{tabular}{llrrr}
        \toprule
        Task & Dataset & Train & Valid & Test \\
        \midrule
        \multirow{3}{*}{\textbf{\negprobing{}}
        }
        & SNLI & 25104 & 484 & 456 \\
        & MNLI & 126232 & 3180 & 3246 \\
        & FEVER & 19874 & 2180 & \multicolumn{1}{c}{--} \\
        \midrule
        \multirow{2}{*}{\textbf{\lexprobing{}}
        }
        & SNLI & 35388 & 734 & 732 \\
        & MNLI & 18542 & 518 & 464 \\
        \midrule
        \multirow{2}{*}{\textbf{\subprobingcond{}.}
        }
        & SNLI & 4438 & 234 & 226 \\
        & MNLI & 5432 & 202 & 154 \\
        \bottomrule
    \end{tabular}
    \caption{Number of samples in all probing datasets created from the different base datasets.}
    \label{tab:stats_probing_tasks}
\end{table}

\subsection{Evaluation}
\label{sec:mdl}
We use a linear probe across all experiments. 
We evaluate both the probe's accuracy and its minimum description length (MDL;  \citealt{voita2020informationtheoretic}), to measure bias extractability. 
Formally, given a dataset $\mathcal{D} =\left\{\left(x_1,y_1\right),\ldots,\left(x_n,y_n\right)\right\}$ and a probabilistic model $p_\theta \left( y \mid x \right)$,
the description length of the model is defined as the number of bits required to transmit the labels $Y = \left( y_1, \ldots, y_n \right)$, given $X = \left( x_1, \ldots, x_n \right)$.
We estimate MDL using \citeauthor{voita2020informationtheoretic}'s \textit{online coding}, and denote the result %
$\Lon$. Given a uniform distribution over the $K$ labels, we get $\Lunif = \lvert \mathcal{D} \rvert \log K$. Thus, the \textit{compression} is defined as $\mathcal{C} = \frac{\Lunif}{\Lon}$ and it holds that $1 \leq \mathcal{C} \leq \mathcal{C}^*$ where $\mathcal{C}^*$ is the compression given by a perfect model.
We interpret a lower MDL score (and consequently, a higher compression score) to mean that the probing property is more extractable from the model representation. The hyperparameters we use in the evaluation process are outlined in Appendix~\ref{sec:appendix:mdl}.

\subsection{Debiasing Methods}

To deploy our framework in the context of robustness to bias, we examine several proposed strategies for debiasing NLU models. In all cases, a weak learner models the bias and is combined with a main model to produce less biased predictions. 

We note that there are three different criteria for controlling the debiasing strategy: (1) Models may be trained  \emph{end-to-end} by propagating errors to the weak learner as well as the main model \citep{mahabadi2020endtoend} or in a \emph{pipeline}, where the weak learner is trained first and frozen, such that only its predictions are used to tune the combination loss   \citep{he2019unlearn, clark2019dont, sanh2020learning,UtamaDebias2020degrading}. (2) The bias model can accept the bias either \emph{explicitly} (by accepting only a set of predefined biased features $x^b$, as in most work) or \emph{implicitly}, by training it in a weak setting: \citet{sanh2020learning} train a small model (TinyBERT; \citealt{turc2019wellread}) and rely on its limited size to adopt biased representations, while \citet{utama2020debiasing} train a BERT-size model on a small subset of the training set, to allow it to capture weaker features of the data. (3) The \emph{objective function} by which the main and bias model are combined can vary. Below we describe three common objective functions. We test different combinations of all strategies where they are feasible, resulting in a wide array of debiased models. 

\subsubsection{Debiasing Objectives}
\paragraph{Debiased Focal Loss (DFL)} 
Focal loss was first proposed by \citet{lin2018focal} to encourage a classifier to focus on the harder examples, for which the model is less confident. This is achieved by weighing standard cross-entropy with $\left( 1 - p_m \right)^\gamma$, where $p_m$ is the class probability and $\gamma$ is the focusing parameter. \citet{mahabadi2020endtoend} propose DFL, where the weighting is achieved by a bias-only model's class probability $p_b$ and the loss becomes:
\begin{equation}
\label{eqn:dfl}
    - \frac{1}{N} \sum_{i=1}^N { (
        1 - p_b
        )^\gamma
        \log {
            p_m
        }
    }.
\end{equation}
We re-implement their model with two bias-only models: a hypothesis-only model and a lexical bias model that uses the same input features as \citet{mahabadi2020endtoend}, outlined in Appendix~\ref{sec:appendix:dfl}

\paragraph{Product of Experts (PoE)}
Product of experts (PoE) was first proposed by \citet{Hinton00trainingproducts} as a method for training ensembles of models that are experts at specific sub-spaces of the entire distribution space. Each model can focus on an ``area of expertise'' and their multiplied predictions form the combination model. This idea was utilized in several studies \citep{he2019unlearn,clark2019dont,mahabadi2020endtoend, sanh2020learning} to train a combination of models where the experts are weak models. The combination model output becomes
\begin{equation}
\begin{split}
    \mathcal{F}_c \left( x \right) &= \mathrm{softmax} \left( 
        \log p_b + 
        \log p_m
    \right),
\end{split}
\end{equation}
and is trained with standard cross-entropy.

\paragraph{Confidence Regularization (ConfReg)}
In this method, proposed by \citet{UtamaDebias2020degrading}, a bias-only/weak model and a teacher model are first independently trained on the target dataset. Then, the predictions of the teacher model are down-weighed by the predictions of the weak model. The weighted loss is then used to distill knowledge \citep{hinton2015distilling} to a new main model, parameterized in the same way as the teacher model (this is known as \emph{self distillaion}).
We note that ConfReg cannot be easily trained in an end-to-end setting, because it relies on an already trained teacher model to down-weigh the predictions.

\section{Experiments}
\label{sec:exps}
\subsection{Datasets}

\begin{table*}[t]
    \centering
    \begin{tabular}{p{1cm}p{1.25cm}rrrrrr}
\toprule
                && \multicolumn{3}{c}{\textbf{\lexprobing}} & \multicolumn{3}{c}{\textbf{\subprobing}} \\
                \cmidrule(lr){3-5} 
                \cmidrule(lr){6-8}
                 Bias & Model &
                 \multicolumn{1}{c}{$\mathcal{C}$} & 
                 \multicolumn{1}{c}{Acc.} & $\mathrm{HANS}^-$ & 
                 \multicolumn{1}{c}{$\mathcal{C}$} & 
                 \multicolumn{1}{c}{Acc.} & $\mathrm{HANS}^-$ \\
\midrule
                 & Random &                     $1.4 \pm 0.0$ &           $59.7 \pm 2.7$ & \multicolumn{1}{c}{--} &                     $1.4 \pm 0.0$ &           $64.9 \pm 4.5$ &                 \multicolumn{1}{c}{--} \\
                 & Pretrained &                     $1.9 \pm 0.0$ &           $77.4 \pm 0.2$ & \multicolumn{1}{c}{--} &                     $1.9 \pm 0.0$ &           $80.9 \pm 0.6$ &                 \multicolumn{1}{c}{--} \\
                 & Base &                     $3.2 \pm 0.2$ &           $88.8 \pm 1.2$ &   $38.9 \pm 18.8$ &                     $3.2 \pm 0.3$ &           $91.3 \pm 3.6$ &     $6.5 \pm 3.0$ \\
                 \midrule
\multirow{2}{*}{Explicit}
                & DFL\etoe &                     $4.0 \pm 0.5$ &           $92.6 \pm 1.3$ &    $67.4 \pm 9.7$ &                     $4.4 \pm 0.5$ &           $95.1 \pm 2.6$ &    $28.4 \pm 6.6$ \\
                 & PoE\etoe &                     $4.0 \pm 0.5$ &           $91.7 \pm 0.7$ &    $65.3 \pm 4.8$ &                     $4.2 \pm 0.5$ &           $92.5 \pm 0.7$ &    $17.4 \pm 1.8$ \\
                 \midrule
\multirow{2}{*}{Subset}
                & ConfReg &                     $4.6 \pm 0.5$ &           $92.3 \pm 1.6$ &   $53.2 \pm 14.2$ &                     $4.3 \pm 0.4$ &           $93.4 \pm 1.7$ &    $18.4 \pm 5.9$ \\
                 & DFL &                     $4.1 \pm 0.1$ &           $92.2 \pm 0.7$ &   $57.1 \pm 13.0$ &                     $3.9 \pm 0.2$ &           $93.5 \pm 2.0$ &   $38.4 \pm 16.4$ \\
                 \midrule
\multirow{4}{*}{Tiny}
                & DFL &                     $4.8 \pm 0.3$ &           $\mathbf{93.6 \pm 1.1}$ &    $\mathbf{75.3 \pm 4.8}$ &                     $4.6 \pm 0.4$ &           $94.7 \pm 1.9$ &    $45.9 \pm 6.9$ \\
                 & DFL\etoe &                     $\mathbf{4.9 \pm 0.3}$ &           $93.0 \pm 1.0$ &    $74.0 \pm 5.8$ &                     $\mathbf{4.7 \pm 0.2}$ &           $\mathbf{95.1 \pm 1.4}$ &    $\mathbf{57.6 \pm 9.6}$ \\
                 & PoE &                     $3.6 \pm 0.3$ &           $90.9 \pm 1.1$ &    $63.5 \pm 5.5$ &                     $3.9 \pm 0.5$ &           $93.3 \pm 1.3$ &    $13.2 \pm 4.2$ \\
                 & PoE\etoe &                     $4.2 \pm 0.1$ &           $92.0 \pm 0.8$ &    $73.1 \pm 6.6$ &                     $4.3 \pm 0.2$ &           $94.3 \pm 2.3$ &    $27.2 \pm 5.2$ \\
\bottomrule
\end{tabular}
    \caption{Results of probing for \lexprobing{} and \subprobing{} on MNLI. $\mathcal{C}$ is the compression of the probing classifier and \emph{Acc} is the accuracy. $\mathrm{HANS}^-$ identifies the performance of the original model on the relevant subset of non-entailed samples in HANS: (1) the lexical overlap subset for \textbf{\lexprobing}, (2) the subsequence subset for \textbf{\subprobing}. We report results for models with different bias models: (1) explicit bias-only model with lexical overlap features, (2) implicit bias model with subsampling (Subset), and (3) implicit TinyBERT bias model (Tiny).}
    \label{tab:results_lex_mnli}
\end{table*}

We use three English NLU datasets:  SNLI, MNLI and FEVER.
 They are used both for training baseline and debiased models, and to create probing datasets for our tasks, as described in Section~\ref{sec:data_processing}.\footnote{Our probing tasks contain examples from all original labels of the datasets. A reviewer pointed out that one can look at probing datasets where examples are drawn only from a specific down-stream label, but our experiments found that splitting per label does not reveal different trends than those we observe here.}

\paragraph{SNLI}
The SNLI dataset contains around 570k premise-hypothesis pairs with three possible labels: \textbf{entailment} if the premise entails the hypothesis, \textbf{contradiction} if the premise contradicts the hypothesis, or \textbf{neutral} if neither hold.
We evaluate on the hard subset \citep{gururangan-etal-2018-annotation}, designed to have fewer hypothesis-only biases. 
\paragraph{MNLI}
The MNLI dataset is a multi-genre variant of SNLI which contains around 430k premise-hypothesis pairs. We evaluate on a hard subset of the dev matched set, provided by \citet{mahabadi2020endtoend}, which was created by taking examples that a hypothesis-only classifier failed to classify. 

\paragraph{FEVER}
The Fact Extraction and VERification (FEVER) dataset contains around 180k pairs of claim--evidence pairs, where the task is to predict one of three labels: either the evidence \textbf{supports} or \textbf{refutes} the claim, or there is \textbf{not enough information}. We evaluate on \textbf{FEVER-Symmetric}, which was designed such that it cannot be predicted by a claim-only classifier~\citep{schuster2019debiasing}.

\subsection{Models}

We test different models based on BERT, by removing the classification head and using the pooled representation of the \texttt{[CLS]} token as input to our probes. 
In settings where previous work compared in-distribution and o.o.d performance, we use hyperparameters which are known to work well for the task and dataset. For new settings which were not reported in previous work, we sweep for the best hyperparameters based on the in-distribution accuracy on the validation set.\footnote{In our experiments, some methods did not converge, notably PoE and DFL using a model with subset sampling. This method was used to train ConfReg models and is likely much more sensitive to selection of the weak model.}
All hyperparameters are available in Appendix~\ref{sec:appendix:hyperparams}. 
We train all models with five random seeds and report means and standard deviations, to account for known variability of fine-tuned models, espeically when evaluated out of distribution \citep{mccoy-etal-2020-berts}.

We reimplement all debiasing methods in a unified codebase to facilitate a fair comparison. Training details are available in Appendix~\ref{sec:appendix:training}.

\paragraph{Baselines}
We use the standard base BERT implementation of \citet{wolf2020huggingfaces}. We take the pretrained model without further fine-tuning on any downstream task (denoted as Pretrained) and we also fine-tune the model on the target dataset (Base). To obtain a lower bound on the performance of these models, we take the same model and randomly initialize its weights (Random).

\section{Results}
\label{sec:results}
In this section, we first report our main finding---the correlation between the robustness of models.
We then analyze each bias type and dataset in a more fine-grained manner.

Table~\ref{tab:correlations} shows the Pearson correlations ($\rho$) between robustness and bias extractability. Robustness is measured as the difference between the performance of a debiased model on a relevant o.o.d dataset and that of a baseline model. Higher values mean that the debiased model is more robust.
Bias extractability is measured as the compression score using a probing classifier designed to target the bias. 
In all but one case, we find positive correlations, indicating that the more successful a method is in debiasing model predictions, the more it makes the bias accessible in the inner representations. 

The only exception is \negprobing{} bias on MNLI, where we report a negative correlation. As we analyze below, in this case some models do not improve on o.o.d data, but their compression still increases. This suggests that even though various debiasing methods are not always successful on different datasets and bias types, they still make bias more accessible in the representations.

\begin{table}[ht]
    \centering
    \begin{tabular}{llcr}
        \toprule
        Bias & Dataset & \multicolumn{1}{c}{$M$} & \multicolumn{1}{c}{$\rho$}  \\
        \midrule
        \multirow{3}{*}{\textbf{\negprobing{}} 
        }
        & SNLI & 6 & $0.757$ \\
        & MNLI & 7 & $-0.257$ \\
        & FEVER & 7 & $0.289$ \\
        \midrule
        \multirow{2}{*}{\textbf{\lexprobing{}} 
        }
        & SNLI & 7 & $0.752$ \\
        & MNLI & 8 & $0.358$  \\
        \midrule
        \multirow{2}{*}{\textbf{\subprobingcond{}.} 
        }
        & SNLI & 7 & $0.672$ \\
        & MNLI & 8 & $0.626$ \\
        \bottomrule
    \end{tabular}
    \caption{Correlation between bias extractability and robustness in various bias types and datasets. $M$ = number of models over which the correlation is measured.}
    \label{tab:correlations}
    \vspace{-10pt}
\end{table}

\begin{table*}[ht]
    \centering
    \begin{tabular}{p{1cm}p{1.25cm}rrrrrr}
\toprule
                 && \multicolumn{3}{c}{\textbf{\lexprobing}} & \multicolumn{3}{c}{\textbf{\subprobing}} \\
                \cmidrule(lr){3-5} 
                \cmidrule(lr){6-8}
                 Bias & Model &
                 \multicolumn{1}{c}{$\mathcal{C}$} & 
                 \multicolumn{1}{c}{Acc.} & $\mathrm{HANS}^-$ & 
                 \multicolumn{1}{c}{$\mathcal{C}$} & 
                 \multicolumn{1}{c}{Acc.} & $\mathrm{HANS}^-$ \\
\midrule
                 & Random &                   $1.4 \pm 0.0$ &         $61.1 \pm 2.1$ & \multicolumn{1}{c}{--} &                   $1.4 \pm 0.0$ &         $55.8 \pm 3.9$ & \multicolumn{1}{c}{--} \\
                 & Pretrained &                   $2.2 \pm 0.0$ &         $83.0 \pm 0.1$ & \multicolumn{1}{c}{--} &                   $2.2 \pm 0.0$ &         $81.2 \pm 0.2$ & \multicolumn{1}{c}{--} \\
                 & Base &                   $4.6 \pm 0.4$ &         $93.8 \pm 1.1$ &   $48.4 \pm 6.3$ &                   $5.4 \pm 0.9$ &         $94.7 \pm 2.3$ &    $2.4 \pm 1.1$ \\
                 \midrule
\multirow{2}{*}{Explicit} 
                & DFL\etoe &                   $\mathbf{5.8 \pm 0.6}$ &         $\mathbf{94.6 \pm 0.3}$ &   $69.1 \pm 9.7$ &                   $\mathbf{6.7 \pm 0.8}$ &         $\mathbf{95.2 \pm 2.1}$ &  $\mathbf{21.0 \pm 18.9}$ \\
                 & PoE\etoe &                   $4.9 \pm 0.3$ &         $93.8 \pm 0.7$ &  $65.0 \pm 10.9$ &                   $5.7 \pm 0.5$ &         $95.0 \pm 0.9$ &    $7.9 \pm 4.4$ \\
                 \midrule
\multirow{1}{*}{Subset} 
                & ConfReg &                   $4.2 \pm 0.3$ &         $93.4 \pm 0.6$ &  $62.0 \pm 10.3$ &                   $4.4 \pm 0.4$ &         $93.0 \pm 0.9$ &   $14.9 \pm 6.0$ \\
                \midrule
\multirow{4}{*}{Tiny} 
                & DFL &                   $4.1 \pm 0.5$ &         $92.6 \pm 1.6$ &   $55.7 \pm 8.7$ &                   $4.5 \pm 1.0$ &         $91.8 \pm 2.2$ &    $6.9 \pm 4.3$ \\
                 & DFL\etoe &                   $5.0 \pm 0.3$ &         $94.2 \pm 0.7$ &   $69.4 \pm 8.2$ &                   $5.6 \pm 0.6$ &         $94.7 \pm 1.5$ &   $13.6 \pm 6.7$ \\
                 & PoE &                   $5.0 \pm 0.4$ &         $93.9 \pm 0.7$ &   $64.6 \pm 9.3$ &                   $6.0 \pm 0.6$ &         $94.5 \pm 0.9$ &   $13.5 \pm 4.8$ \\
                 & PoE\etoe &                   $4.9 \pm 0.3$ &         $94.2 \pm 0.4$ &   $\mathbf{70.8 \pm 5.1}$ &                   $5.7 \pm 0.7$ &         $94.7 \pm 1.6$ &   $15.6 \pm 6.8$ \\
\bottomrule
\end{tabular}
    \caption{Results of probing for lexical bias on SNLI. The notation here stays consistent with Table~\ref{tab:results_lex_mnli}.}
    \label{tab:results_lex_snli} 
\end{table*}

\subsection{Lexical Bias}

\paragraph{MNLI}
Table \ref{tab:results_lex_mnli} shows results for the \lexprobing{}/\subprobing{} probing tasks, on MNLI. For each model, we report compression ($\mathcal{C}$) and accuracy of the probe and the performance of the model on anti-biased (non-entailed) samples from the relevant subset of HANS attributed to the lexical overlap heuristic ($\mathrm{HANS}^-$ column). 

All debiasing methods improve the o.o.d generalization (performance on $\mathrm{HANS}^-$) compared to the base model, consistent with prior work. 
All debiasing methods also lead to models with more extractable bias, as demonstrated by higher compression values. 
The base model already exhibits higher compression than a random model or a pre-trained model, indicating that fine-tuning makes bias more extractable from the inner representation. However, fine-tuning with any debiasing method makes this bias even more extractable. 

In fact, as performance on the anti-biased examples from the HANS subset increases, so does the compression of the probe; Figure~\ref{fig:mnli_lex_class_scatter} shows an example of this trend in the subsequence case. DFL with implicit bias from the TinyBERT model (trained either end-to-end or in a pipeline) has the highest compression values, as well as the biggest improvement out of distribution.

\paragraph{SNLI}
Table \ref{tab:results_lex_snli} shows results for the \lexprobing{} and \subprobing{} probing tasks. All debiasing methods lead to improved o.o.d performance, as expected. Compression of the random and pretrained baselines remains very close, with most of the bias being made more extractable in the representations of the fine-tuned baseline (Base). Most of the debiased models still largely surpass the baseline for compression and probing accuracy, indicating that they make bias more extractable. ConfReg and DFL with a fine-tuned TinyBERT are exceptions; they do not exhibit higher compression than the baseline, but still improve out of distribution. 

\subsection{Negative Word Bias}

\begin{table}[t]
    \centering
    \begin{tabular}{p{0.9cm}p{1.25cm}rHr}
\toprule
                 Bias & Model & \multicolumn{1}{c}{$\mathcal{C}$} &&    Symmetric \\
\midrule
                 & Random &                   $1.37 \pm 0.00$ &         $56.9 \pm 1.3$ &                 \multicolumn{1}{c}{--} \\
                 & Pretrained &                   $1.64 \pm 0.03$ &         $71.0 \pm 0.1$ &                 \multicolumn{1}{c}{--} \\
                 & Base &                   $2.97 \pm 0.10$ &         $85.0 \pm 1.7$ &  $56.0 \pm 2.0$ \\
                 \midrule
\multirow{2}{*}{Claim}
                & DFL\etoe &                   $3.04 \pm 0.08$ &         $87.6 \pm 1.2$ &  $62.1 \pm 1.8$ \\
                 & PoE\etoe &                   $3.00 \pm 0.05$ &         $\mathbf{87.9 \pm 0.5}$ &  $61.9 \pm 1.6$ \\
                 \midrule
\multirow{1}{*}{Subset} 
                & ConfReg &                   $3.03 \pm 0.04$ &         $87.5 \pm 1.3$ &  $56.2 \pm 2.0$ \\
                \midrule
\multirow{4}{*}{Tiny}
                & DFL &                   $\mathbf{3.31 \pm 0.09}$ &         $87.7 \pm 0.7$ &  $\mathbf{62.2 \pm 3.9}$ \\
                 & DFL\etoe &                   $3.16 \pm 0.07$ &         $86.9 \pm 1.0$ &  $60.5 \pm 2.5$ \\
                 & PoE &                   $3.12 \pm 0.05$ &         $87.5 \pm 0.8$ &  $61.0 \pm 3.6$ \\
                 & PoE\etoe &                   $3.06 \pm 0.06$ &         $86.4 \pm 1.0$ &  $61.4 \pm 3.2$ \\
\bottomrule
\end{tabular}
    \caption{Results of probing for \textbf{NegWords} on FEVER. Symmetric is the o.o.d set by  \citet{schuster2019debiasing}, which is designed such that a claim-only classifier cannot achieve higher-than-guess performance on it. Probing accuracy is reported in Appendix~\ref{sec:appendix:results}.}
    \label{tab:results_fever}
\end{table}

\paragraph{FEVER}
Table \ref{tab:results_fever} shows the results for the NegWords task on FEVER. All models improve on FEVER-Symmetric compared to the baseline (Base), indicating that they are less biased in their predictions. Conversely, when probed for the bias, all models achieve higher compression compared to the baseline and outperform it in terms of probing accuracy. That is, this bias is more extractable in the debiased models than in the baseline model.
As a point of reference, the compression of the random model is smallest, closely followed by the pre-trained model. Any fine-tuning leads to significantly larger compression scores. These trends are consistent with the \lexprobing{}/\subprobingcond{}.\ results.
The best model in terms of o.o.d accuracy is DFL with an implicit TinyBERT bias model. We also see that bias is most extractable in this model, compared to the baseline. While previous work used statistical tools to show that the \textsc{Refutes} label is spuriously correlated with negative bigrams \citep{schuster2019debiasing}, we reveal that this information is preserved and even amplified in the model when an attempt is made to make the predictions less reliant on it.

\paragraph{SNLI}
In this case, all models perform better or as well as the baseline model when evaluated on the hard subset, yet the compression values of all models significantly surpass the baseline. While any form of debiasing makes bias more available in the representations, it does not necessarily lead to an improvement on the o.o.d set. Models with a hypothesis-only model perform best out of distribution, and also expose the most bias. Similarly to the results on FEVER, the compression of the random and pretrained models is significantly lower and close to each other, with most of the bias being made available by fine-tuning the model (Base).
Table~\ref{tab:results_snli_neg} in Appendix~\ref{sec:appendix:results} provides the full results.

\paragraph{MNLI}
Compression results are much closer to the fine-tuned baseline, but all debiased models still contain more information about negation words. This is on-par with previous results that anaylzed the statistical correlation of such negation words to the \textsc{contradiction} label \citep{gururangan-etal-2018-annotation, poliak-etal-2018-hypothesis}, and we show that not only does the correlation exist in the data, but attempts to remove such evidence result in more extractability. Still, most growth in compression compared to the random and pretrained models is attributed to the fine-tuning process itself (without debiasing). Interestingly, some of the models do not improve the performance on the hard test set, but their compression still increases, suggesting that the more accessible bias can also be decoupled from the predictions of the model. Table~\ref{tab:results_mnli_neg} in  Appendix~\ref{sec:appendix:results} provides the full results. 

\subsection{Varying the Debiasing Effect}
\begin{figure}[ht]
    \centering
    \includegraphics[width=\linewidth]{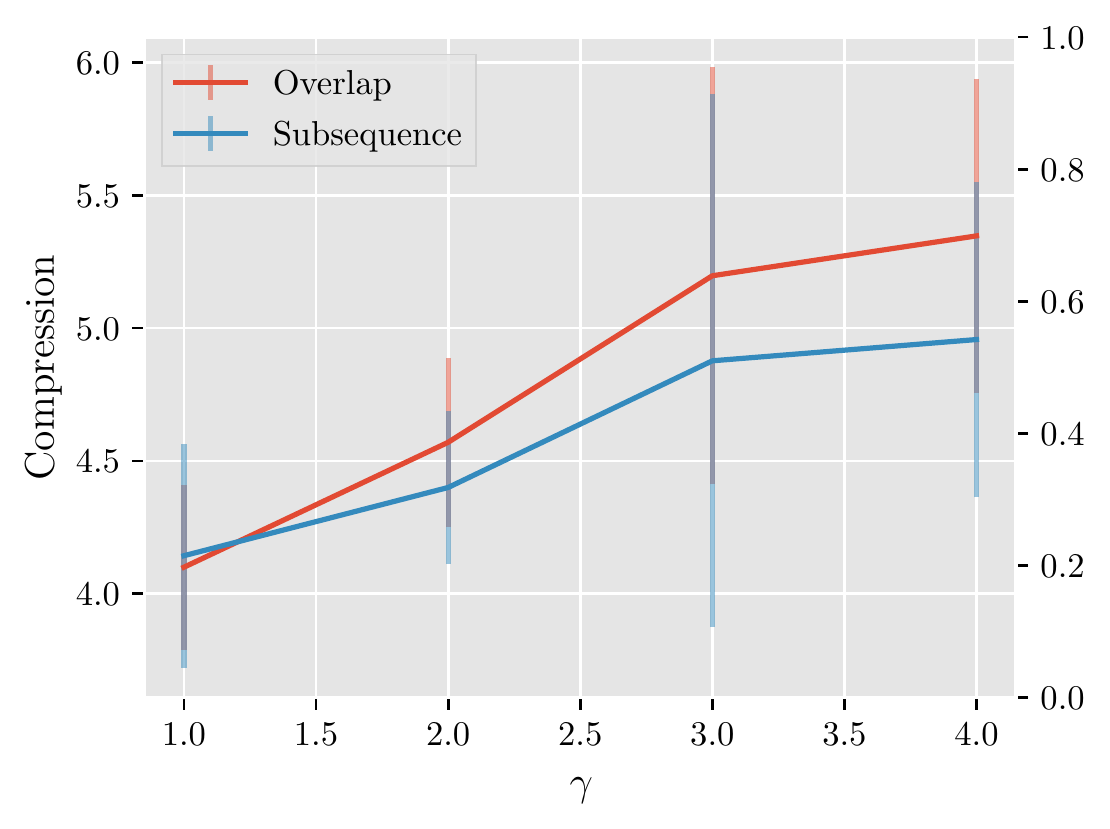}
    \caption{Compression of a DFL model with an implicit bias model on \lexprobing{}/\subprobing{} probing tasks vs.\ the focusing parameter $\gamma$, for MNLI. As $\gamma$ increases, the bias becomes more extractable.}
    \label{fig:exp_varying_gamma}
\end{figure}

So far we evaluated the effect of debiasing on bias extractability across debiasing methods. 
To evaluate this effect within the same method, we analyze the effect of stronger debiasing in the DFL method, by increasing the ``focusing parameter'' $\gamma$ (Eq.~\ref{eqn:dfl}). 
 We test our probing tasks on models trained with increasing values of $\gamma \in \left\{ 1, 2, 3, 4 \right\}$. 
 Figure \ref{fig:exp_varying_gamma} shows the results for the \lexprobing{}/\subprobing{} tasks. As we increase $\gamma$, the extractability of bias from the model's representations increases. This is consistent with our main results.

\subsection{Linguistic Information in Debiased Models}

\begin{table*}[ht]
    \centering
    \begin{tabular}{llllll}
\toprule
{} &           
\multicolumn{4}{c}{$\mathcal{C}$} \\
\cmidrule(lr){2-5}
& 
\multicolumn{1}{c}{Random} &      
\multicolumn{1}{c}{Pretrained} &         
\multicolumn{1}{c}{Baseline} & 
\multicolumn{1}{c}{Average} &            
\multicolumn{1}{c}{Accuracy} \\
\midrule
BShift    &  $1.39 \pm 0.01$ &  $2.41 \pm 0.0$ &  $1.61 \pm 0.01$ &       $1.67 \pm 0.03$ &  $51.6 \pm 0.57$ \\
CoordInv  &  $1.37 \pm 0.01$ &  $1.58 \pm 0.0$ &  $1.48 \pm 0.01$ &        $1.5 \pm 0.02$ &  $59.0 \pm 1.74$ \\
ObjNum    &   $1.4 \pm 0.01$ &  $1.79 \pm 0.0$ &  $1.77 \pm 0.02$ &       $1.86 \pm 0.04$ &   $73.8 \pm 1.0$ \\
SOMO      &   $1.37 \pm 0.0$ &  $1.48 \pm 0.0$ &  $1.44 \pm 0.01$ &       $1.45 \pm 0.01$ &   $58.7 \pm 0.5$ \\
Tense     &  $1.48 \pm 0.01$ &  $3.05 \pm 0.0$ &  $2.38 \pm 0.12$ &        $2.52 \pm 0.1$ &  $83.8 \pm 1.25$ \\
SentLen   &   $3.0 \pm 0.16$ &  $2.19 \pm 0.0$ &   $1.49 \pm 1.2$ &       $2.24 \pm 0.06$ &   $50.8 \pm 0.8$ \\
SubjNum   &   $1.39 \pm 0.0$ &  $2.11 \pm 0.0$ &  $1.83 \pm 0.03$ &       $1.96 \pm 0.05$ &  $76.2 \pm 0.92$ \\
TopConst  &   $1.68 \pm 0.0$ &  $2.82 \pm 0.0$ &  $2.41 \pm 0.06$ &       $2.47 \pm 0.14$ &  $51.9 \pm 3.28$ \\
TreeDepth &   $1.48 \pm 0.0$ &  $1.55 \pm 0.0$ &  $1.53 \pm 0.01$ &       $1.56 \pm 0.01$ &   $25.6 \pm 0.6$ \\
\bottomrule
\end{tabular}
    \caption{Average accuracy and compression scores for debiased models and baselines, when probed for the SentEval tasks \citep{conneau-etal-2018-cram}. Random is the randomly initialized model, Pretrained is the pretrained model without fine-tuning, and Base is the fine-tuned model. Accuracy and Average denote the average accuracy and compression score of $M$ debiased models trained on MNLI ($M = 8$). }
    \label{tab:sent_eval}
\end{table*}

Following the main results, a useful question to ask is whether debiased models also tend to learn useful linguistic information more broadly, which may explain the noticeable increase in performance out of distribution.\footnote{
We thank a reviewer for pointing out this idea.
} To test this, we take our models trained for NLI on the MNLI dataset and apply the SentEval probing tasks \citep{conneau-etal-2018-cram}, which test ten different linguistic properties in model representations. We exclude the word content (WC) task, because it is a 1000-way classification problem and takes substantially more time to train with an MDL probe. Table~\ref{tab:sent_eval} shows the average results for all debiased models and the remaining nine tasks, compared to our three baselines (random, pretrained, fine-tuned). First, we notice that for 8/9 tasks, compression of the model decreases when it is fine-tuned, compared to the pretrained model. This can be explained by the close connection between the linguistic phenomena and the masked language modelling (MLM) objective, compared to fine-tuning on NLI. Furthermore, on average, debiased models do not decrease in compression compared to the fine-tuned model, but the differences are very subtle and generally within standard deviation bounds. This suggests that while debiasing does not make linguistic information measured in these probing tasks less extractable, it also does not substantially amplify it, as opposed to extractability of bias information.

\section{Discussion and Conclusion}
\label{sec:discussion}

All of our experiments tested model-based debiasing, where a weak learner is used to capture biased features and discourage their use in model predictions. We discover that for both explicit and implicit modeling of the bias, this method exposes the biased features in the representation. When we fix the model and change the effect of debiasing (through the ``focusing parameter'' of DFL), we observe the same trend, where stronger bias mitigation leads to higher extractability of the modelled bias. Based on our results, we stipulate that while current debiasing methods are good at making model predictions less biased, they are a bad proxy for learning unbiased text representations. The increased extractability of bias from the representations is not necessarily a bad trait: For example, the \negprobing{} task does not reveal more granular semantics of negation, which may be useful for the generalization of the model. By probing for linguistic properties using the SentEval tasks, we also observe that debiased models do not make linguistic information less extractable, which can also contribute to their improvement in performance. We argue that future research should look for more interpretable methods for debiasing language models, and consider the problem of finding robust, bias-free feature detectors.

Another domain where this finding may be alarming is social bias. Previous studies show that word vectors contain social bias \citep{Caliskan2017SemanticsDA}, and that debiasing them does not necessarily remove this information \citep{gonen-goldberg-2019-lipstick}. Our work shows that debiasing sometimes increases the information available about bias in the representations, albeit in the context of dataset bias rather than social bias. 

Our work shows that \emph{unbiased predictions} $\implies$ \emph{biased representations}. We speculate that there exists a proxy for the language model that removes bias information from the representations and consequently improves the generalization of predictions out of distribution. Future work could focus on methods that are both representation-robust and prediction-robust w.r.t various biases. Finding such methods can help alleviate leakage of bias from data to the model's representations, without sacrificing the in-distribution performance.

\section*{Acknowledgments}
\label{sec:acks}
This research was supported by the ISRAEL SCIENCE FOUNDATION (grant No. 448/20) and by an Azrieli Foundation Early Career Faculty Fellowship. We thank Victor Sanh and Prasetya Ajie Utama for their feedback and help reproducing the models. We also thank the anonymous reviewers for their useful suggestions. 

\bibliography{bibliography}
\bibliographystyle{acl_natbib}

\clearpage
\appendix

\section{Appendix}
\label{sec:appendix}
\subsection{Online Code Evaluation}
\label{sec:appendix:mdl}

Following \citet{voita2020informationtheoretic}, we evaluate our models using an online code probe, with timestamps [2.0, 3.0, 4.4, 6.5, 9.5, 14.0, 21.0, 31.0, 45.7, 67.6, 100] (Each timestamp corresponds to a percentage of the samples in the training dataset). We use a slightly different scale than \citet{voita2020informationtheoretic}, to account for the smaller datasets and the resulting instability in the first fractions of training. The last timestamp is used to train the probe on the full training dataset, and it is then evaluated for accuracy on the entire test set. During all training phases, we employ early-stopping when the validation accuracy does not improve over four epochs, with a tolerance of $10^{-3}$.

\subsection{Bias-only Models}
\label{sec:appendix:dfl}
For the lexical bias-only model, we use the following features as bias input features: 1) Whether all words in the hypothesis are included in the premise; 2) If the hypothesis is the contiguous subsequence of the premise; 3) If the hypothesis is a subtree in the premise’s parse tree; 4) The number of tokens shared between premise and hypothesis normalized by the number of tokens in the premise, and 5) The cosine similarity between premise and hypothesis’s pooled token representations from BERT followed by min, mean, and max-pooling. Following \citeauthor{mahabadi2020endtoend}, we also give equal weights to neutral and contradiction labels (by calculating a weighted cross-entropy loss) to encourage the model towards biased predictions.

\subsection{Hyperparameters}
\label{sec:appendix:hyperparams}

\paragraph{ConfReg}
We train all models for five epochs and use the same hyperparameters as in \citet{utama2020debiasing}: 2000 samples for the weak learner sub-sampling, a batch size of 32, learning rate of \numprint{5e-5}, a weight decay of 0.01 and a linear scheduler for modulating the learning rate with a 10\% warm-up proportion. For training FEVER, we set a learning rate of \numprint{2e-5} and sub-sample 500 samples. For SNLI we use the same parameters as in MNLI, but we sub-sample \numprint{3000} samples to account for the larger dataset, and make sure that the weak model still follows the constraints: at least 90\% of the predictions on the sampled training set fall within the $0.9$ probability bin, and the weak learner achieves more than 60\% accuracy on the entire training set.

\paragraph{DFL and PoE}
We train all models for three epochs on MNLI and SNLI with a batch size of 32, learning rate of \numprint{5e-5}, a weight decay of 0.0 and a linear scheduler for modulating the learning rate with a 10\% warm-up proportion. We choose $\gamma = 2.0$ for most of the DFL models. Exceptions are made for DFL with the subsampled bias model and end-to-end DFL with a TinyBERT bias model, where we sweep $\gamma \in \left\{ 1.0, 2.0 \right\}$ and choose $\gamma = 1.0$ based on the highest validation accuracy (in-distribution). Another exception is made for FEVER, where we set the learning rate at \numprint{2e-5} to be consistent with previous work.

\subsection{Training Details}
\label{sec:appendix:training}

To train all models, we have used single instances of NVIDIA GeForce RTX 2080 Ti, with an average training time of 1–7 hours. Models where the weak learner is frozen have 110M parameters, as in the base BERT model. TinyBERT models have 4.4M parameters \citep{turc2019wellread} and any combination of a weak model and a main model is straightforward to calculate.

\subsection{Additional Results}
\label{sec:appendix:results}
Table~\ref{tab:results_snli_neg} summarizes the results for \negprobing{} bias on the SNLI dataset, and Table~\ref{tab:results_mnli_neg} summarizes the results on MNLI. Table~\ref{tab:results_fever_appendix} shows the full results for \negprobing{} on FEVER, including probing accuracy.

\begin{table*}[t]
    \centering
    \ra{1.1}
    \begin{tabular}{llrrr}
\toprule
                 Bias & Model & \multicolumn{1}{c}{$\mathcal{C}$} & \multicolumn{1}{c}{Acc.} &              \multicolumn{1}{c}{Hard} \\
\midrule
                 & Random &                   $1.47 \pm 0.0$ &         $59.8 \pm 2.4$ & \multicolumn{1}{c}{--} \\
                 & Pretrained &                   $2.01 \pm 0.0$ &         $76.1 \pm 0.0$ & \multicolumn{1}{c}{--} \\
                 & Base &                   $3.48 \pm 0.3$ &         $92.9 \pm 0.4$ &  $80.51 \pm 0.57$ \\
                 \midrule
\multirow{2}{*}{Hypothesis}
                & DFL\etoe &                   $5.24 \pm 0.3$ &         $95.4 \pm 0.7$ &  $82.91 \pm 0.38$ \\
                 & PoE\etoe &                   $5.23 \pm 0.2$ &         $95.9 \pm 0.4$ &  $82.37 \pm 0.46$ \\
                 \midrule
\multirow{4}{*}{Tiny}
                & DFL &                   $5.13 \pm 0.3$ &         $95.6 \pm 0.9$ &  $80.5 \pm 0.9$ \\
                 & DFL\etoe &                   $4.49 \pm 0.6$ &         $94.1 \pm 0.9$ &  $80.06 \pm 0.62$ \\
                 & PoE &                   $4.81 \pm 0.2$ &         $94.0 \pm 0.8$ &  $81.4 \pm 0.4$ \\
                 & PoE\etoe &                   $4.41 \pm 0.4$ &         $94.3 \pm 0.8$ &  $80.4 \pm 0.3$ \\
\bottomrule
\end{tabular}
    \caption{Results of probing for \textbf{\negprobing{}} on SNLI. We also report results on the SNLI hard test set from \citet{gururangan-etal-2018-annotation}}
    \label{tab:results_snli_neg}
\end{table*}

\begin{table*}[t]
    \centering
    \ra{1.1}
    \begin{tabular}{llrrr}
\toprule
                 Bias & Model & \multicolumn{1}{c}{$\mathcal{C}$} & \multicolumn{1}{c}{Acc.} & \multicolumn{1}{c}{Hard} \\
                 \midrule
                 & Random &                   $1.48 \pm 0.01$ &         $56.8 \pm 0.57$ & \multicolumn{1}{c}{--} \\
                 & Pretrained &                   $1.57 \pm 0.00$ &         $52.8 \pm 0.0$ & \multicolumn{1}{c}{--} \\
                 & Base &                   $2.42 \pm 0.11$ &         $85.2 \pm 1.1$ &  $76.7 \pm 0.2$ \\
                 \midrule
\multirow{2}{*}{Hypothesis}
                & DFL\etoe &                   $2.66 \pm 0.11$ &         $86.5 \pm 0.4$ &  $77.8 \pm 0.9$ \\
                 & PoE\etoe &                   $2.60 \pm 0.06$ &         $86.1 \pm 1.4$ &  $77.4 \pm 0.5$ \\
                 \midrule
\multirow{1}{*}{Subset}
                & ConfReg &                   $2.85 \pm 0.07$ &         $88.2 \pm 0.3$ &  $76.6 \pm 0.5$ \\
                 \midrule
\multirow{4}{*}{Tiny}
                & DFL &                   $2.75 \pm 0.06$ &         $88.4 \pm 0.1$ &  $76.5 \pm 0.0$ \\
                 & DFL\etoe &                   $2.68 \pm 0.10$ &         $87.5 \pm 1.0$ &  $75.6 \pm 0.4$ \\
                 & PoE &                   $2.71 \pm 0.23$ &         $87.4 \pm 1.3$ &  $77.8 \pm 0.9$ \\
                 & PoE\etoe &                   $2.64 \pm 0.09$ &         $87.5 \pm 0.1$ &  $76.8 \pm 0.1$ \\
\bottomrule
\end{tabular}
    \caption{Results of probing for \textbf{\negprobing{}} on MNLI. We also report results on the MNLI hard test set generated by \citet{mahabadi2020endtoend}}
    \label{tab:results_mnli_neg}
\end{table*}

\begin{table*}[t]
    \centering
    \begin{tabular}{p{0.9cm}p{1.25cm}rrr}
\toprule
                 Bias & Model & \multicolumn{1}{c}{$\mathcal{C}$} & Acc. &    Symmetric \\
\midrule
                 & Random &                   $1.37 \pm 0.00$ &         $56.9 \pm 1.3$ &                 \multicolumn{1}{c}{--} \\
                 & Pretrained &                   $1.64 \pm 0.03$ &         $71.0 \pm 0.1$ &                 \multicolumn{1}{c}{--} \\
                 & Base &                   $2.97 \pm 0.10$ &         $85.0 \pm 1.7$ &  $56.0 \pm 2.0$ \\
                 \midrule
\multirow{2}{*}{Claim}
                & DFL\etoe &                   $3.04 \pm 0.08$ &         $87.6 \pm 1.2$ &  $62.1 \pm 1.8$ \\
                 & PoE\etoe &                   $3.00 \pm 0.05$ &         $\mathbf{87.9 \pm 0.5}$ &  $61.9 \pm 1.6$ \\
                 \midrule
\multirow{1}{*}{Subset} 
                & ConfReg &                   $3.03 \pm 0.04$ &         $87.5 \pm 1.3$ &  $56.2 \pm 2.0$ \\
                \midrule
\multirow{4}{*}{Tiny}
                & DFL &                   $\mathbf{3.31 \pm 0.09}$ &         $87.7 \pm 0.7$ &  $\mathbf{62.2 \pm 3.9}$ \\
                 & DFL\etoe &                   $3.16 \pm 0.07$ &         $86.9 \pm 1.0$ &  $60.5 \pm 2.5$ \\
                 & PoE &                   $3.12 \pm 0.05$ &         $87.5 \pm 0.8$ &  $61.0 \pm 3.6$ \\
                 & PoE\etoe &                   $3.06 \pm 0.06$ &         $86.4 \pm 1.0$ &  $61.4 \pm 3.2$ \\
\bottomrule
\end{tabular}
    \caption{Results of probing for \textbf{NegWords} on FEVER, including probe accuracy (Acc.).}
    \label{tab:results_fever_appendix}
\end{table*}

\end{document}